\documentclass[11pt,twocolumn,letterpaper]{article}

\usepackage{iccv}
\usepackage{times}
\usepackage{epsfig}
\usepackage{graphicx}
\usepackage{amssymb}
\usepackage{algorithm}
\usepackage{algorithmic}
\usepackage{bbm}
\usepackage{amsfonts}
\usepackage{mathrsfs}
\usepackage{multirow}
\usepackage{amsmath}
\usepackage{paralist}
\usepackage{xcolor}
\usepackage[pagebackref=true,breaklinks=true,colorlinks,bookmarks=false]{hyperref}

\iccvfinalcopy % *** Uncomment this line for the final submission

\def\iccvPaperID{406} % *** Enter the CVPR Paper ID here

\begin{document}

%%%%%%%%% TITLE
\title{On the application of transfer learning in prognostics and health management}

% Authors List
\author{Ramin Moradi{$^{1}$},~ Katrina M. Groth{$^{1}$}\vspace{-8pt}{\small~}\\
\\
{$^{1}$}Center for Risk and Reliability, University of Maryland, MD, USA\\
{\small{Corresponding Author: \tt{raminmrd@umd.edu}}}
}

% Create the title
\maketitle

% Abstract
\begin{abstract}
Advancements in sensing and computing technologies, the development of human and computer interaction frameworks, big data storage capabilities, and the emergence of cloud storage and could computing have resulted in an abundance of data in modern industry. This data availability has encouraged researchers and industry practitioners to rely on data-based machine learning, specially deep learning, models for fault diagnostics and prognostics more than ever. These models provide unique advantages, however their performance is heavily dependent on the training data and how well that data represents the test data. This issue mandates fine-tuning and even training the models from scratch when there is a slight change in operating conditions or equipment. Transfer learning is an approach that can remedy this issue by keeping portions of what is learned from previous training and transferring them to the new application. In this paper, a unified definition for transfer learning and its different types is provided, Prognostics and Health Management (PHM) studies that have used transfer learning are reviewed in detail, and finally a discussion on TL application considerations and gaps is provided for improving the applicability of transfer learning in PHM.    
\end{abstract}

\section{Introduction}

Technology advancements in different fronts including sensing and measurement, data collection practices, data processing and computation, data storage capability, and the emergence of external processing and storage power (i.e. cloud service providers) are all accelerating the transformation of industry to the concept of a data-driven industry 4.0 \cite{lasi2014industry}. Along with this transition, data-driven methods, specially deep learning methods, for PHM applications such as fault detection, diagnostics, and prognostics have attracted great interest. This is due to their unique ability to handle large amounts of data. In fact, the larger the data size, the better their performance. Also, deep learning models are able to automatically generate good-enough features in cases where there is a lack of understanding about the domain for feature engineering. 

However, there are still some issues with these models as well. One of their key problems is the lack of generalizability of a trained model to other equipment, settings and operating conditions in the context of PHM. Transfer Learning (TL) is an approach that can remedy the generalizability issue by storing the knowledge gained while solving one problem and transferring it to a different but related problem. The \textit{Neural Information Processing Systems (NIPS)} 1995 workshop \cite{thrun1998learning} is believed to be the starting point for research on this topic. Since then, terms such as Learning to Learn, Knowledge Consolidation or transfer, inductive transfer, and domain adaptation have been used to convey TL.

In the PHM domain, TL could considerably save great endeavors to manually labeling data and retuning models for new problems. Specially, given the fact that high-quality labeled data that includes failures and is also publicly available is hard to find. Therefore, it is highly desirable to be able to use a model that is trained with a good dataset on a specific equipment and working condition on other related problems. One could imagine several valuable applications for this possibility, such as:
\begin{compactitem}
\item Transferring the knowledge gained by training on one equipment to a fleet of identical equipment with slightly different operation conditions \cite{wang2019domain}.
\item Transferring the knowledge gained from training a diagnostics model using laboratory test data \cite{yang2019intelligent} or simulation data \cite{xu2019digital} to health management of the same equipment in the field.
\item Transferring the knowledge from other domains such as image processing to PHM applications including image-based structural health monitoring \cite{gao2018deep} or fault diagnosis of time series data that is converted to image \cite{wen2019transfer}. 
\end{compactitem}

Despite the above-mentioned possible applications, this approach is not fully embraced by the PHM community (there are only two published TL related papers in \textit{reliability engineering and systems safety} journal and the identified references are dispersed across various journals and conferences). For this reason, in this paper, we provide a formal definition of the TL and its different categories, review the related studies in order to provide a better perspective on the applications of TL for PHM researchers, and finally we discuss the required considerations for using this method.

\section{Transfer Learning}

To mathematically express TL, a domain can be defined as follows: \[D = \{ \mathcal{X}, P(X) \}\] 
Where \(\mathcal{X}\) is a feature space and $P(X)$ is a marginal probability distribution in which \(X = \{x_1, x_2,...,x_n\} \in \mathcal{X} \). 

For a specific domain, a learning task can be defined as: 
\[ \mathcal{T} = \{\mathcal{Y}, f(.)\}\]
Where \(\mathcal{Y}\) is the label  space and \(f(.)\) is the desired predictive function. This function is learned from the training data that is pairs of \(\{x_i, y_i\}\) where \(x_i \in X\) and \( y_i \in \mathcal{Y} \). In probabilistic terms, $f(x)$ can be written as \(P(y|x)\). For any new instance $x$, $f(x)$ would predict the corresponding prediction (e.g. a label in a classification scenario).  

In TL studies usually two domains are considered which are the source domain ($D_S$) and the target domain ($D_T$) (note that there can be several source domains (Multi-Source TL \cite{ge2014handling}). Given a $D_S$, a source learning task \(\mathcal{T}_S\), a $D_T$, and a corresponding target learning task \(\mathcal{T}_T\), the objective of TL is to improve the learning of the target domain's predictive function $f_T(.)$ using the information gained from $D_S$ and $T_S$ where the source and target domain or tasks are not the same.  

In this paper, the categorization introduced by \cite{pan2009survey} is used to classify TL. This categorization has three main criteria which are similarity of source and target domains, similarity of source and target tasks, and availability of labeled data in the source and target domains. In this categorization, TL has three main classes: 

\begin{enumerate}
    \item \textbf{Inductive TL}: the target task is different from the source task (\(\mathcal{T}_S\ \neq \mathcal{T}_T\)) while the source and target domains can be either different or the same. In Inductive TL, label data is available in the target domain but not necessarily in the source domain. In this type of TL, \(\mathcal{T}_S\) and \(\mathcal{T}_T\) can either be learned at the same time (i.e. multi-task learning) or sequentially. 
    
    \item \textbf{Transductive TL}: the source and domain tasks are the same, while there is a domain shift or a distribution change between the source and the target (\(D_S \neq D_T \), \(\mathcal{T}_S = \mathcal{T}_T\)). In this type of TL, labeled data is only available in the source domain. Transductive TL is also called "Domain Adaptation" in many studies. 
    
    \item \textbf{Unsupervised TL}: the source and domain tasks are different but related (similar to inductive TL) and there is no labeled data available in the source or target domains. 
\end{enumerate}

Also, in each class, there are four common TL approaches that can be applied based on "what to transfer" \cite{pan2009survey,weiss2016survey,yan2019knowledge}: 

\begin{compactitem}
\item Parameter-based TL:  transfers knowledge through shared parameters between source and target domain learner models. 
\item Instance-based TL: instances from the source domain are reweighted to compensate for marginal distribution differences between the two domains. The reweighted instances are then directly used for training in the target domain.  
\item Feature-based TL: transfers features from the source domain either by reweighting to better match the target domain features or by discovering a common latent feature space that has acceptable predictive qualities and minimizes the marginal distribution between the domains. 
\item Relevance-based TL: transfers knowledge based on some defined relationship between the two domains.
\end{compactitem}

\section{Transfer Learning in PHM}

In this section, the existing PHM-related studies that have used any one of the three mentioned TL classes are reviewed. 

\subsection{Inductive TL (\(\mathcal{T}_S \neq \mathcal{T}_T\))} 
One of the most studied applications of TL is computer vision and visual classification problems, especially using deep Convolutional Neural Networks (CNN). All of the identified inductive TL studies have used images and CNN in their approaches. For example, \cite{gao2018deep} have used inductive TL for training deep CNN for image-based structural damage recognition. In this study, the low-level feature extractors from the VGG-16 \cite{simonyan2014very} model that is trained using the ImageNet dataset \cite{deng2009imagenet} for classifying different structures (walls, bridges, buildings,etc.) are transferred to the relatively smaller target domain (structural ImageNet) to discern structures with spalling from the healthy ones. 

\begin{figure}[!ht] 
\centering
 \includegraphics[width=0.48\textwidth]{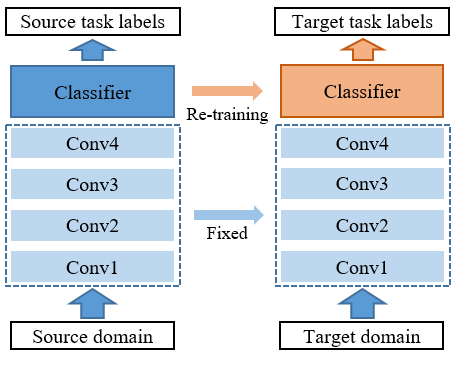}
 \caption{ The general idea of the identified inductive TL. The feature extractor layers are froze and the target domain classifier is re-trained.}
 \label{fig:inductiveTL}
\end{figure}

Using the same source domain and same trained deep CNN model (VGG-16), \cite{shao2018highly} have transferred the the low-level feature extractors of the VGG-16 to a new deep learning model that its task is machine fault diagnosis and condition monitoring. The original sensor data in the target domain are in time series format. The authors have used time-frequency imaging to convert the sensor data to images and increase the similarity between the source and target domains. Following the same approach, \cite{wen2019transfer} has used the ResNet-50 (pre-trained on ImageNet dataset) deep CNN network \cite{he2016deep} as the feature extractor for a machine fault classification network. They too, have converted time series sensor data to RGB images in order to make it compatible with the ResNet-50 inputs. 

\cite{zhong2019novel} proposed training a CNN-based anomaly detector (only two classes: normal and abnormal) on a gas turbine dataset. Then transferring the convolution layers to the target domain classifier (parameter-based TL) and feeding those features to a Support Vector Machine (SVM) for classifying the gas turbine condition into four different classes (1 normal and 3 fault classes). Their approach has significantly improved the fault diagnosis performance with an small amount of labeled data in the target domain.

\subsection{Transductive TL (\(D_S \neq D_T \), \(\mathcal{T}_S = \mathcal{T}_T\)) } 

Transductive TL or domain adaptation is the most widely used type of TL for PHM applications, specially fault diagnostics in the source and target domains. Domain generalization is the main application of TL in PHM that have been identified in the literature. In this application, one general model that is applicable to the source and target domains is produced and labeled source dataset is available while the target dataset is unlabeled. The source domain information is used to predict accurate labels for the target dataset. \cite{qian2019novel} have proposed using Kullback-Leibler (KL) divergence as a criteria to measure the discrepancy between the source and target domain datasets. They have defined a loss function that is the sum of first to $nth$ order moment discrepancies between the two domains. This term is fused into the objective function of the simultaneous training of the source and target fault diagnosis networks to be minimized. This way, the networks would learn to find domain independent features that helps learning the classification task in target domain using the labels in the source domain.   

In another study by \cite{tong2018bearing} on domain generalization for bearing fault diagnosis, in order to reduce the marginal distribution difference between the domains and extract maximally domain-invariant features, Maximum Mean Discrepancy (MMD) measure is used. They use this measure to regularize the two dataset in a way that minimizes the MMD measure. Afterwards, first the classifier is trained on the transformed source domain (labeled). Then, pseudo-labels are generated for the target domain data (unlabeled) which is simply picking up the class that has the maximum predicted probability, as if they were true labels \cite{lee2013pseudo}. Having labels for both domains, the difference between the class-conditional probability distributions can be calculated (using modified MMD measure) and be incorporated into the model training loss function to be minimized and provide a domain-invariant feature generator. To address the same problem as defined by \cite{tong2018bearing}, \cite{sun2019sparse} has used the MMD to measure the difference between the source and target domains hierarchically-obtained features in a Sparse stacked denoising autoencoder architecture. This difference terms are then summed and fused into the training loss function to be minimized and help obtaining domain-invariant features. 

\begin{figure*}[!ht]
\centering
 \includegraphics[width=0.95\textwidth]{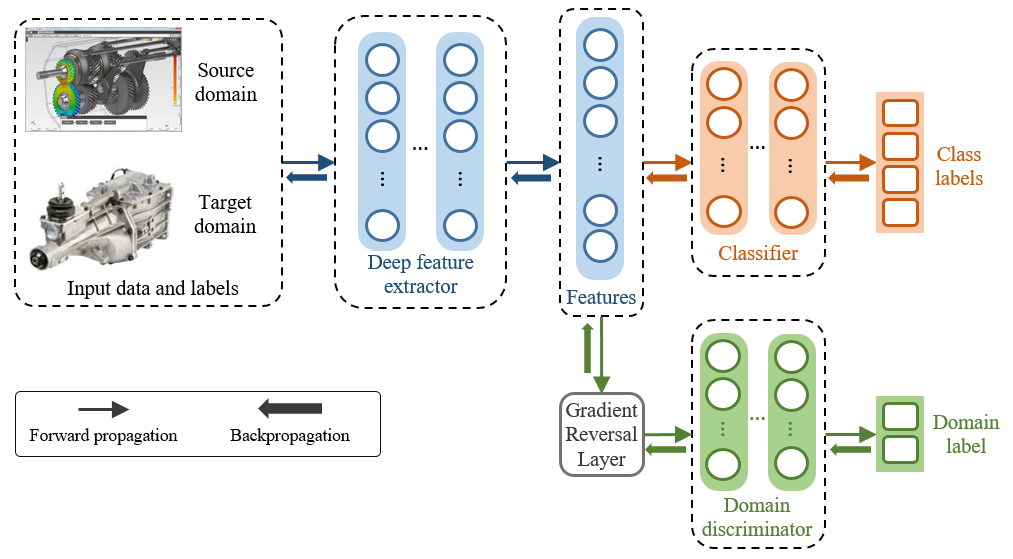}
 \caption{A general representation of DANN implementation for domain adaptation on a hypothetical gearbox simulation-based source domain and real-world target domain. In this architecture, Gradient Reversal Layer (GRL) has no associated parameters. During the forward propagation, the GRL acts as an identity transformation and in the backpropagation it multiplies
 the gradient by a certain negative constant. }
 \label{fig:DANN}
\end{figure*}

\cite{li2020domain,wang2019domain} proposed using a deep learning architecture that is called Domain Adversarial Neural Network (DANN) \cite{ganin2016domain} and includes a deep feature extractor ($G$), a domain discriminator ($D$), and a classifier ($C$) for obtaining a domain-invariant fault classifier for rotating machinery. In this architecture, training data from several different domains as well as augmented data is used. Further, through adversarial training of the deep learning architecture as shown in Figure \ref{fig:DANN} (asking D to identify whether the features are coming from source or target domain while asking G to fool the discriminator), the feature extractor is trained to generate features from data in different domains that are not domain specific. In other words, the machinery data under different working conditions are mapped onto a learned common feature space that is able to classify faults in various domains. 

Considering the explained architecture and TL techniques one can imagine applying TL to prognostics and Remaining Useful Life (RUL) prediction as well. As demonstrated by \cite{zhang2018transfer}, a predictive deep learning model such as a Long Short Term Memory (LSTM) network can be trained for a turbofan engine RUL prediction (using C-MAPSS datasets) in one operating condition and then be transferred to another operation condition using the same architecture that is shown in Figure \ref{fig:inductiveTL} (some parameters are kept and some are retrained). Also, using the same dataset, \cite{da2020remaining} has proposed training a LSTM model for RUL prediction of turbofans with a subtle difference that no labels are used for training in the target domain. To classify the unlabeled target domain, LSTM-DANN is architecture is proposed which is built upon the explain DANN architecture in Figure \ref{fig:DANN}. 

Another interesting applications of transductive TL in PHM is transferring the knowledge gained from simulations and experiments ($D_S$) to real-world problems ($D_T$). One of the growing virtual sources of knowledge are digital twins. Digital twins are a determining technology for the Industrial Internet of Things (IIOT) \cite{canedo2016industrial} where machines can interact with each other and humans in the virtual space. Digital twin includes the virtual and physical spaces as well as the interactions between the two. It models the physical twin in terms of geometry, behaviors, and governing rules. Also, theoretically it can mirror, predict, and verify the performance of the physical entity. 

As shown by \cite{xu2019digital}, using a digital twin, it is possible to generate data (source domain) with an acceptable volume and variety for training a proper initial deep learning-based fault diagnostics model. They have developed such model by training a Stacked Sparse Auto Encoder (SSAE) on data from the digital twin of a car body-side production line and transferred the obtained model parameters (parameter-based TL) to be fine tuned using physical monitoring dataset. Other types of simulation have also been used in the literature as the source domain. To mention a few, using feature-based TL, \cite{li2020transfer} have transferred fault diagnosis knowledge from simulation data of a continuously stirred tank reactor and the pulp mill plant benchmark problem to real-world data. \cite{wang2018distribution} have used a portion of the Tennessee Eastman (TE) process simulation data as the labeled source dataset and the remainder as the unlabeled target dataset. Pseudo-labeling technique and adversarial training between the classifier and domain discriminator is used in this study as well. Considering the experimental data as the source domain and following a similar technical approach, \cite{yang2019intelligent} have transferred locomotive bearing fault diagnostics knowledge to unlabeled real-world operation data. 

\subsection{Unsupervised TL}

The studies that dealt with training diagnostics and prognostics models in cases where there are labeled data in both the source and target domains or only in the source domain are discussed. In our context, collecting the fault data is difficult, specially when it comes to expensive and safety-critical systems. Some machines and systems cannot run to failure due to he associated expenses and/or consequences. Also, even regardless of the expenses and consequences, many industrial systems go through a long degradation process to reach failure which makes the failure data collection a very time consuming task. Therefore, it is desired to still be able to improve the deep learning-based diagnostics and prognostics models performance using the knowledge embedded in other related domains. Unsupervised learning in general includes tasks such as density estimation and clustering, anomaly detection or one-class classification, and learning the latent representation of variables (feature space). 

\begin{figure}[!ht] 
\centering
\includegraphics[width=0.5\textwidth]{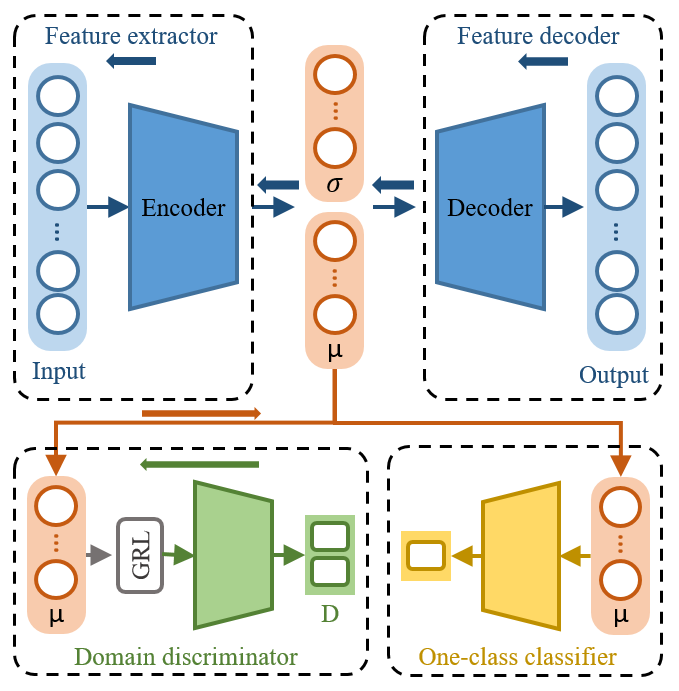}
\caption{The representation of an architecture used in (Michau \& Fink, 2019) for unsupervised TL. The training data from both domains is fed to a variational auto encoder, a domain discriminator is trained in an adversarial manner to force the features to become as domain-invariant as possible, and finally the one-class classifier does the prediction using the generated features (no backpropagation arrow is used in the one-class classifier module). }
\label{fig:unsupervisedTL}
\end{figure}

Accordingly, unsupervised TL uses strategies that assume a known format of transformations between the domains, the availability of discriminative domain-invariant features, a latent space where the difference in distribution of source and target data is minimal, or the possibility of transforming and mapping the source data onto the target domain \cite{gopalan2013unsupervised}. Among the PHM studies, only one study by \cite{michau2019domain} is found that has applied a fully unsupervised TL to learn to detect anomalies in the early life of gas turbines using healthy data operation data of other similar turbines. Their proposed framework is composed of a feature extractor and a one-class classifier trained with only healthy data. To align the features of the source and target domains, they have selected three strategies and explored various combinations of them. A variational autoencoder to generate a shared probabilistic encoder/decoder for the two domains, a homothety loss (a transformation of space which dilates distances with respect to a fixed point) to maintain inter-point spacial relationships between the input data and the extracted features, and an adversarial training of a domain discriminator (as explained in Section 3.2). Their proposed architecture is demonstrated in Figure \ref{fig:unsupervisedTL}. 

\section{Discussion}
TL has significant advantages and applications in the PHM context. However, similar to any other method it has drawbacks and understanding these drawbacks is vital for its successful application and implementation. There are three main questions that should be considered before applying TL which we discuss separately in the following subsections.  

\subsection{When to transfer?}
Transferring knowledge is possible only when it is 'appropriate'. Defining what 'appropriate' means in each context is an ongoing research question and usually careful experimentation is required. First, the need should be identified for TL. For example, when we talk about TL for fleets of systems, we should first identify which units require additional data or knowledge for accurate diagnosis, then determine which other systems in the fleet would be most helpful (for a pairwise TL) and have minimal negative transfer effect. One could consider all the available data on the whole fleet as the source domain, however in this scenario a careful data selection/filtration should be performed to avoid biasing the model with noises and irrelevant information. 

Regarding the amount of data required for TL, usually TL is best applied when the source domain is considerably larger than the target domain. Occasionally, TL can be helpful when the two domains have almost the same size as well. Another important consideration is that if the predictive model has a high prediction error on the source data (due to noisy data, etc.), similar or even worse prediction errors on the transformed target data are expected. Thus we should have confidence in the source domain prediction model before the transfer process. 

\subsection{What to transfer?}

Another important aspect of TL is what is being transferred. As mentioned four main types of information can be transferred from the source to the target domain which are instances, features, shared parameters, and defined relationship between the source and target domains. All these forms of information can be transferred in inductive TL, while in transductive TL only the instances and features can be transferred, and in unsupervised TL the only the features can be transferred.  

\subsection{How to transfer?}

Various approaches are discussed throughout this paper that are not perfect and have some inherent flaws and weaknesses. For examples, the use of similarity metrics between distributions (i.e. MMD, KL, etc.) was explained. It is shown that this measures are not always reliable and there is a need for more robust measures \cite{zhang2019transfer}. We mentioned pseudo-labeling as a way to improve the performance of the models in the target domain (unlabeled), however incorrect pseudo-labels can considerably worsen the performance of the model and they should be used with caution. Also even with pseudo-labeling, in cases with unlabeled target domains, the confidence in class predictions becomes very low if the disparity between the domains is large. 

Adversarial training is growing popular in machine learning applications and it is being more and more used in TL applications. As mentioned a feature generator and a domain discriminator are the common elements of adversarial training for TL. The domain discriminator's only duty is to distinguish the source and domain features and it can easily overfit the data. Therefore, it cannot consider task-specific decision boundaries between classes, leading to ambiguous features near class boundaries which means less robust generated features. 

\section{Conclusion and future work }

TL has become a major field of study in machine learning. It mimics an important feature of human learning and improves the machine learning performance. For the PHM society, TL can be a way to reduce the required training data and significantly improve fault diagnostics and prognostics ability in situations with limited data and information. In this study we thoroughly reviewed the studies that used TL for PHM purposes. The low number of identified studies and the fact that most of them are published in the last couple of years, shows the increasing adoption of this technique by PHM researchers. Thus, this paper could contribute to the field by introducing the different state of the art techniques, identified applications, and implementation considerations. 

Future studies could fruitfully explore TL application in the PHM domain further by enabling transfer of knowledge between more diverse tasks, such as fault diagnosis of similar systems with manufacturers and different condition monitoring configurations. Also, transferring knowledge from simulation-based source domains needs further exploration. Exciting emerging technologies like digital twins could potentially be the perfect source domain for target industrial systems. 

{\small
\bibliographystyle{ieee_fullname}
\bibliography{main.bib}
}

\end{document}